\pdfoutput=1

\documentclass[11pt]{article}

\usepackage{acl}

\usepackage{times}
\usepackage{latexsym}
\usepackage{comment}
\usepackage{mathtools}
\usepackage[T1]{fontenc}
\usepackage{graphicx}
\usepackage{tikz}
\usepackage{pgfplots}
\usepackage{multirow}
\usepackage{subcaption}
\usepackage{float}
\pgfplotsset{compat = 1.3}
\usepackage{hyperref}

\usepackage[utf8]{inputenc}

\usepackage{microtype}

\title{Contrastive Visual Semantic Pretraining Magnifies the Semantics of Natural Language Representations}

\author{Robert Wolfe \\
  University of Washington \\
  \texttt{rwolfe3@uw.edu} \\\And
  Aylin Caliskan \\
  University of Washington \\
  \texttt{aylin@uw.edu} \\}

\begin{document}
\maketitle
\begin{abstract}
We examine the effects of contrastive visual semantic pretraining by comparing the geometry and semantic properties of contextualized English language representations formed by GPT-2 and CLIP, a zero-shot multimodal image classifier which adapts the GPT-2 architecture to encode image captions. We find that contrastive visual semantic pretraining significantly mitigates the anisotropy found in contextualized word embeddings from GPT-2, such that the intra-layer self-similarity (mean pairwise cosine similarity) of CLIP word embeddings is under $.25$ in all layers, compared to greater than $.95$ in the top layer of GPT-2. CLIP word embeddings outperform GPT-2 on word-level semantic intrinsic evaluation tasks, and achieve a new corpus-based state of the art for the RG65 evaluation, at $.88$. CLIP also forms fine-grained semantic representations of sentences, and obtains Spearman's $\rho = .73$ on the SemEval-2017 Semantic Textual Similarity Benchmark with no fine-tuning, compared to no greater than $\rho = .45$ in any layer of GPT-2. Finally, intra-layer self-similarity of CLIP sentence embeddings decreases as the layer index increases, finishing at $.25$ in the top layer, while the self-similarity of GPT-2 sentence embeddings formed using the EOS token increases layer-over-layer and never falls below $.97$. Our results indicate that high anisotropy is not an inevitable consequence of contextualization, and that visual semantic pretraining is beneficial not only for ordering visual representations, but also for encoding useful semantic representations of language, both on the word level and the sentence level.
\end{abstract}

\section{Introduction}

Large-scale "natural language supervision" using image captions collected from the internet has enabled the first "zero-shot" artificial intelligence (AI) image classifiers, which allow users to create their own image classes using natural language, yet outperform supervised models on common language-and-image tasks \cite{radford2021learning}. The image encoders of such models have been shown to form "multimodal" representations in the upper layers, such that the same neurons fire for photographic, symbolic, and textual depictions of a concept \cite{goh2021multimodal}. Research on these state of the art "visual semantic" (joint language-and-image) models has focused primarily on their benefits for encoding semantically legible representations of images. In this paper, we seek to answer a straightforward but as yet unexplored question: what benefits does contrastive visual semantic pretraining have for representations of natural language?

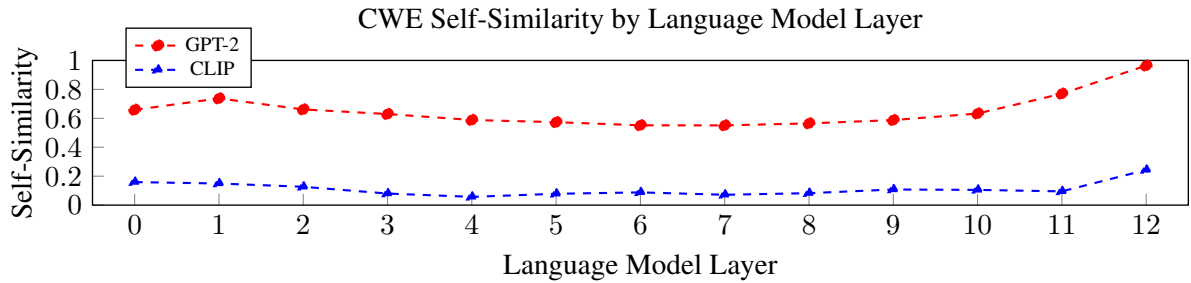
\begin{figure*}[htbp]
\begin{tikzpicture}
\begin{axis} [
    height=3.5cm,
    width=\linewidth,
    line width = .5pt,
    ymin = 0, 
    ymax = 1,
    xmin=-.5,
    xmax=12.5,
    ylabel=Self-Similarity,
    ylabel shift=-5pt,
    xtick = {0,1,2,3,4,5,6,7,8,9,10,11,12},
    xtick pos=left,
    ytick pos = left,
    title=CWE Self-Similarity by Language Model Layer,
    xlabel= {Language Model Layer},
    legend style={at={(.03,.81)},anchor=south west,nodes={scale=0.7, transform shape}}
]
\addplot[thick,dashed,mark=*,color=red] coordinates {(0,0.6587458800100342) (1,0.7372493841393961) (2,0.6612104705462435) (3,0.6294690014865569) (4,0.5884662538197166) (5,0.5731922100920905) (6,0.5517696641206877) (7,0.5511352858413449) (8,0.5647982609283869) (9,0.5877102865850451) (10,0.6330118142510739) (11,0.7698747042012253) (12,0.964203893021691)};

\addplot[thick,dashed,mark=triangle*,color=blue] coordinates {(0,0.15913507178224345) (1,0.14935120094529203) (2,0.1266165790034515) (3,0.07992260756237189) (4,0.05707932176630996) (5,0.07830196844146199) (6,0.08802649450408537) (7,0.07145548666837015) (8,0.08260096449145693) (9,0.10793742483538417) (10,0.1052025078351228) (11,0.09503851254085677) (12,0.24313851016362073)};

\legend {GPT-2, CLIP};

\end{axis}
\end{tikzpicture}
\caption{CLIP CWEs are much less self-similar than GPT-2 CWEs, despite being trained with the same architecture, suggesting that pretraining objective leads to significant differences in contextualized representations which are not the result of the contextualization process itself, nor of the model architecture.}
\label{selfsim_by_layer}
\end{figure*}

The CLIP ("Contrastive Language Image Pretraining") image classification model introduced by \citet{radford2021learning} provides a unique opportunity to observe the effects of visual semantic pretraining on a contextualizing language model. While most other visual semantic architectures combine language and image features in the inner layers of the model \cite{lu2019vilbert}, CLIP separates the language model from the vision model until the end of the encoding process, at which point it projects a representation formed by each model into a joint language-image embedding space \cite{radford2021learning}. CLIP is trained to maximize the cosine similarity of a projected image with its projected natural language caption, while minimizing the cosine similarity of the projected caption with all of the other images in the batch \cite{radford2021learning}, a training objective known as "contrastive learning" or "contrastive representation distillation" \cite{tian2019contrastive}. The separation of the language model from the vision model prior to projection allows us to consider the two models independently of each other, such that we can study representations of natural language trained for a visual semantic objective, rather than representations which combine language and image features in the inner layers of the model. Moreover, because CLIP encodes natural language using GPT-2, a "causal" language model trained solely on next-word prediction, we can directly compare representations formed using the same architecture, but for two very different objectives: one solely linguistic, the other visual semantic.

We observe differences between representations formed by GPT-2 and the CLIP language model ("LM") both on the word level and on the sentence level. We outline our contributions:

\begin{enumerate}
    \item As shown in Figure \ref{selfsim_by_layer}, contrastive visual semantic pretraining mitigates the angular uniformity (known as anisotropy, measured using cosine similarity) observed by \citet{ethayarajh2019contextual} in GPT-2 and other contextualizing LMs. The intra-layer self-similarity (mean pairwise cosine similarity, where $1.0$ is maximally similar and $0.0$ maximally dissimilar) of contextualized word embeddings (CWEs) is less than .25 in all layers of the CLIP LM, compared to greater than $.50$ in all layers and greater than .95 in the top layer of GPT-2. The five highest-magnitude neuron activations in a CWE from the CLIP LM make up $39$\% of its length in the top layer, compared to more than $97$\% of the length of a top layer GPT-2 CWE. This indicates that high anisotropy is not an inescapable consequence of contextualization, nor of using a specific language modeling architecture, but is dependent on pretraining objective, and is significantly reduced by using an objective which is both contrastive and visual semantic. 
    \item Contrastive visual semantic pretraining results in CWEs which outperform other static and contextualized word embeddings on word-level intrinsic evaluation tasks. CLIP word embeddings obtained in a "decontextualized" setting (wherein the model is given only the word with no other context) set new state of the art for a corpus-based method on the RG65 intrinsic evaluation task \cite{rubenstein1965contextual}, with Spearman's $\rho = .88$ in the eighth layer of the CLIP LM, and match state of the art for the ValNorm task, which evaluates the semantic quality of representations based on correspondence with pleasantness norms \cite{toney2020valnorm}, with Pearson's $\rho = .88$ in layer 4. CLIP CWEs outperform GPT-2 CWEs on every intrinsic evaluation in a decontextualized setting, and for all but one evaluation also outperform the GPT-2 embeddings of \citet{bommasani2020interpreting}, who encode $100,000$ contexts and pool over the representations to form a static word embedding matrix.
    \item Contrastive visual semantic pretraining encodes semantically useful sentence representations which obtain Spearman's $\rho = .73$ on the SemEval-2017 Semantic Textual Similarity (STS) Benchmark using the cosine similarity between sentence pairs. CLIP results on the STS benchmark outperform those of GPT-2, which never exceed $\rho = .45$ in any layer of the model. Moreover, we find that while GPT-2 sentence embeddings formed using the end-of-sequence (EOS) token exhibit intra-layer self-similarity $\geq .97$ in all layers, the self-similarity of CLIP sentence embeddings steadily decreases over the layers of the model, from $.98$ to $.25$ in the top layer, indicating that the contrastive visual semantic pretraining objective of the model forces the formation of fine-grained semantic representations of sentences, such that they can be associated with encoded images.
\end{enumerate}

We make our code and data available at {\small \url{https://github.com/wolferobert3/clip_contrastive_acl_2022}}.

\section{Related Work}

We review prior work on visual semantic AI, on the geometry and semantic properties of representations formed by language models, and on semantic intrinsic evaluation tasks.

\subsection{Foundation Models}

We examine CLIP and GPT-2, both of which are "foundation models," a term coined by \citet{bommasani2021opportunities} to describe the group of architecturally similar state of the art AI systems which have seen wide adoption across domains including language \cite{raffel2020exploring}, vision \cite{dosovitskiy2020image}, medicine \cite{rasmy2021med}, and programming \cite{chen2021evaluating}, and which exhibit unexpected emergent properties such as strong performance on tasks on which they were not explicitly trained \cite{brown2020language}. GPT-2 and CLIP adapt the transformer neural network architecture, which uses an "attention" mechanism to draw information from the most relevant elements in the model's context window \cite{vaswani2017attention}.

\subsection{Contextualizing Language Models}
GPT-2 is a contextualizing language model, meaning that it forms word representations which incorporate information from surrounding words ("context") \cite{radford2019language}. Such representations, referred to as "contextualized word embeddings" \cite{DBLP:journals/corr/abs-1802-05365}, differ depending on the sense of the word used and the specific context in which the word occurs \cite{soler2021let}, allowing such representations to overcome many of the limitations of static word embeddings, which use only one vector to represent each word \cite{collobert2011natural}. GPT-2 is an autoregressive "causal" language model, meaning that it is trained to predict the next word, and employs "masked self-attention," such that the model can only draw information from words which precede the current word \cite{radford2019language}.

\subsection{CLIP and Visual Semantic AI}

CLIP is a "multimodal" model which combines language and image representations in a single joint visual semantic embedding space \cite{radford2021learning}. CLIP can be used with either a ResNet \cite{he2016deep} or a Vision Transformer (ViT) \cite{dosovitskiy2020image} to encode images, and a language model (GPT-2) to encode captions \cite{radford2019language}. CLIP projects the encoded images and captions into a joint embedding space, where the model maximizes the cosine similarity of the correct image-caption pair while minimizing the cosine similarity of each caption with every other image in the batch \cite{radford2021learning}. CLIP projects only a representation of the entire caption into the joint language-image space, and uses CWEs in order to produce this representation.

CLIP is not the first transformer-based model to form visual semantic representations: both \citet{lu2019vilbert} and \citet{li2019visualbert} adapt the BERT language model of \citet{devlin-etal-2019-bert} to produce visual semantic language-image representations, and \citet{zhang2020contrastive} and \citet{jia2021scaling} use the same contrastive loss objective as CLIP. What makes CLIP unique is that it is the first image classifier to generalize to zero-shot image classification, such that users can define image classes "on-the-fly" using natural language, and obtain performance competitive with supervised computer vision models, without ever fine-tuning on the data for a task \cite{radford2021learning}. CLIP improved the zero-shot state-of-the-art\footnote{\citet{tiwary_2021} report that their Turing Bletchley model improves the zero-shot state of the art to $79.0$\%. This model is not available open source to the research community.} on ImageNet \cite{deng2009imagenet} to $76.2$\% \cite{radford2021learning}, from a previous best of $11.5$\% \cite{li2017learning}.

\subsection{Language Model Geometry}
\citet{ethayarajh2019contextual} find that CWEs in ELMo \cite{peters2018deep}, BERT \cite{devlin-etal-2019-bert}, and GPT-2 \cite{radford2019language} are highly anisotropic (angularly uniform, based on measurements of cosine similarity). The effect is most pronounced in GPT-2, such that randomly selected embeddings in the top layer of the model have "nearly perfect" (\textit{i.e.,} close to $1.0$) cosine similarity \cite{ethayarajh2019contextual}. \citet{cai2020isotropy} find that the inner layers of GPT and GPT-2 form contextualized word representations on a swiss-roll manifold, while BERT embeds words in clusters. Mitigating anisotropy has been shown to be beneficial for semantic representations, as \citet{mu2018all} find that increasing the isotropy (angular dispersion) of static word embeddings improves performance on semantic intrinsic evaluation tasks. \citet{voita-etal-2019-bottom} find that the pretraining objective of a contextualizing language model affects what information is encoded in CWEs, and that embeddings in causal language models (like GPT-2) contain less mutual information with the input token and more mutual information with the next token in the sequence as the layer index increases. \citet{tenney2019bert} shows that layers of BERT are devoted primarily to certain natural language processing (NLP) tasks, and that task complexity increases with the layer index.

\subsection{Intrinsic Evaluation Tasks}

Intrinsic evaluation tasks assess the quality of word or sentence embeddings by measuring the correlation of the geometric properties of the embeddings with human-rated judgments of similarity \cite{tsvetkov2016correlation} or psycholinguistic norms \cite{toney2020valnorm}. \citet{bommasani2020interpreting} create static word embeddings by pooling over CWEs derived from tens of thousands of sentences from English Wikipedia, and study the performance of these embeddings on word-level intrinsic evaluation tasks. They find that embeddings from the upper layers of BERT and GPT-2 perform poorly relative to embeddings from earlier layers, and that embeddings formed by pooling over a word's CWEs significantly outperform embeddings formed from "decontextualized" words, input to the model with no surrounding context \cite{bommasani2020interpreting}. We report results on the four intrinsic evaluation tasks analyzed by \citet{bommasani2020interpreting}, as well as the recently introduced ValNorm task \cite{toney2020valnorm}, and a sentence-level intrinsic evaluation task, the Semantic Textual Similarity Benchmark \cite{cer2017semeval}.

\section{Data}

For comparison of our results on CWE anisotropy with the prior work of \citet{ethayarajh2019contextual}, we encode the text of the SemEval Semantic Textual Similarity tasks from 2012 through 2016 \cite{agirre2012semeval,agirre2013sem,agirre2014semeval,agirre2015semeval}, who used these datasets because they include instances of the same words used in different contexts and reflecting different word senses. We discard sentences too long to fit in the 77-token context window of the CLIP LM, which still leaves us with over 36,000 sentences. 

\subsection{Intrinsic Evaluation Tasks}\label{intrinsic_eval_subsection}

We report results on five word-level tasks:
\begin{itemize}
    \item \textbf{RG-65} \cite{rubenstein1965contextual}, a set of $65$ noun pairs assigned scores between 0 and 4 based on their semantic similarity, as judged by $51$ human participants in a controlled psychological study intended to evaluate the relationship between "similarity of context and similarity of meaning."
    \item \textbf{WordSim-353}, a word relatedness task consisting of 353 word pairs divided into two sets \cite{finkelstein2001placing}. WS-353 was introduced in the context of information retrieval for search engines but is now widespread as an evaluation of word relatedness.
    \item \textbf{SimLex-999}, a word similarity task consisting of $666$ noun-noun word pairs, $222$ verb-verb word pairs, and $111$ adjective-adjective word pairs \cite{hill2015simlex}.
    \item \textbf{SimVerb-3500}, a set of $3,500$ verb pairs rated on similarity by $843$ study participants, and designed to remediate the lack of resources for evaluating verb semantics \cite{gerz2016simverb}. 
    \item \textbf{ValNorm}, which measures the quality of an embedding based on how well it reflects the valence norms of the language on which was trained \cite{toney2020valnorm}. ValNorm takes Pearson's correlation coefficient of human ratings in a valence lexicon with Single-Category Word Embedding Association Test (SC-WEAT) \cite{caliskan2017semantics} pleasantness effect sizes for a word embedding.
\end{itemize}

Finally, we report results on a sentence-level task, the \textbf{Semantic Textual Similarity (STS) Benchmark}, a set of $8,628$ sentence pairs derived from SemEval STS tasks between 2012 and 2017 and rated  on similarity \cite{cer2017semeval}. Sentences reflect three genres: news, forums, and captions. The test set, on which we report results without use of the training set, includes $1,379$ sentence pairs.

\subsection{Language Model Architectures}
While the CLIP LM is based on the GPT-2 architecture, there are minor differences between the models we examine.\footnote{We use the PyTorch models available via the Transformers library of \citet{wolf-etal-2020-transformers}.} The CLIP LM is a 63-million parameter version of the GPT-2 architecture, and uses 12 layers to form 512-dimensional CWEs within a 77-token context window \cite{radford2021learning}. GPT-2 Small, the model studied by \citet{ethayarajh2019contextual} and examined in this paper, forms 768-dimensional CWEs over a 1,024-token context window, and has a total parameter count of 124-million \cite{radford2019language}. Though it consists only of image captions, the text component of the WebImageText corpus used to train CLIP has a "similar" word count to the WebText corpus used to train GPT-2, according to \citet{radford2021learning}.

\section{Approach and Experiments}

We outline our experiments, and discuss our approach for extracting both CWEs and sentence embeddings, and for computing self-similarity.

\subsection{Geometry of CWEs}

We use the self-similarity formula of \citet{ethayarajh2019contextual} to study whether the contrastive visual semantic pretraining objective of CLIP has affected the anisotropy of GPT-2 CWEs:

\begin{equation}
    s = \frac{1}{n^2 - n} \sum_{i} \sum_{j \neq i}  cos(\vec{w_i}, \vec{w_j})
\label{self_sim_equation}
\end{equation}

Note that $cos$ in Equation \ref{self_sim_equation} refers to cosine similarity, or the angular similarity of two vectors after normalization to unit length, a common method for measuring the semantic similarity of word embeddings. $n$ refers to the number of word embeddings $w$ used in the self-similarity measurement. Following \citet{guo2021detecting}, who report consistent results on semantic bias analyses by randomly sampling $10,000$ CWEs, we measure the self-similarity of $10,000$ randomly selected CWEs in contexts from the STS 2012-2016 tasks for every layer of CLIP and GPT-2. We collect CWEs for the same $10,000$ word indices from all layers, rather than randomly selecting new words at every layer. 

Because \citet{mu2018all} find that a few high-magnitude dimensions cause anisotropy and distort the semantics of static word embeddings, we also examine whether CLIP embeddings encode less of their magnitude in a few high-value dimensions. \citet{mu2018all} find that there are usually $n/100$ such distorting dimensions in static word embeddings, where $n$ refers to the embedding's dimensionality. Because GPT-2 small forms 768-dimensional embeddings, and CLIP forms 512-dimensional embeddings, we report the mean proportion of magnitude contained in the top $8$ and the top $5$ neuron activations for each model at each layer across $10,000$ embeddings.

\subsection{Word-Level Intrinsic Evaluation Tasks}

We examine the layerwise performance of CWEs extracted from the CLIP LM and from GPT-2 on the five word-level intrinsic evaluation tasks described in Section \ref{intrinsic_eval_subsection}. For these tasks, we extract the vector corresponding to the last subtoken of every word, as prior work finds that the last subtoken in a causal language model fully encodes the semantics of words which a causal language model breaks into subwords \cite{guo2021detecting}. For each task, we input words in the "decontextualized" setting described by \citet{bommasani2020interpreting} (\textit{i.e.,} with no surrounding context). Unlike \citet{bommasani2020interpreting}, we also extract the BOS token and EOS token from the GPT-2 tokenizer, and add them to either side of the decontextualized word. We do this to keep the experiment consistent between the models, as adding the tokens is default behavior for the CLIP LM, but not for GPT-2. Because it is common to omit the BOS and EOS tokens when using GPT-2, we report results for GPT-2 both with the tokens and without them. To observe whether CLIP sentence embeddings have unique properties, since they are the only linguistic representations projected to the joint language-image space, we also report results on these tasks using the EOS token for the CLIP LM and GPT-2. 

\subsection{Sentence-Level Evaluations}

We report layerwise performance using sentence representations obtained from CLIP and GPT-2 on the STS benchmark \cite{cer2017semeval}. For this task, we use the EOS token in both CLIP and in GPT-2. For GPT-2, we also use the last subtoken of the sentence, with no EOS token added.

Finally, we analyze the self-similarity of sentence embeddings from each model using Equation \ref{self_sim_equation}. In this case, $w$ refers not to a word embedding, but to a sentence embedding. For this analysis, we use embeddings of all of the unique sentences in the test set of STS Benchmark \cite{cer2017semeval}.

\begin{figure}[htbp]
\begin{tikzpicture}
\begin{axis} [
    height=4.5cm,
    width=.45\textwidth,
    line width = .5pt,
    ymin = 0, 
    ymax = 100,
    xmin=-.5,
    xmax=12.5,
    ylabel=\% of Magnitude,
    ylabel shift=-5pt,
    xtick = {0,1,2,3,4,5,6,7,8,9,10,11,12},
    xtick pos=left,
    ytick pos = left,
    title=\% Magnitude in Top $n$ Neurons,
    xlabel= {Layer},
    legend style={at={(.15,.62)},anchor=south west,nodes={scale=0.5, transform shape}}
]

\addplot[thick,dashed,mark=*,color=red] coordinates {(0,44.67385411262512) (1,60.5259895324707) (2,54.28526997566223) (3,55.46831488609314) (4,55.34985661506653) (5,56.85415267944336) (6,56.9152295589447) (7,58.65005850791931) (8,62.1255099773407) (9,64.22644257545471) (10,66.09758734703064) (11,71.69216275215149) (12,97.39335775375366)};

\addplot[thick,dashed,mark=square*,color=orange] coordinates {(0,49.29197430610657) (1,62.33704686164856) (2,56.74484372138977) (3,57.577699422836304) (4,57.341378927230835) (5,58.65538716316223) (6,58.750033378601074) (7,60.46096086502075) (8,63.703131675720215) (9,65.61618447303772) (10,67.3029899597168) (11,72.60722517967224) (12,97.4625825881958)};

\addplot[thick,dotted,mark=triangle*,color=blue] coordinates {(0,45.17318606376648) (1,36.329248547554016) (2,35.12235879898071) (3,31.403520703315735) (4,29.755255579948425) (5,30.359065532684326) (6,30.22181987762451) (7,30.567991733551025) (8,31.20342791080475) (9,32.53809213638306) (10,33.14794600009918) (11,30.956920981407166) (12,39.29747939109802)};

\addplot[thick,dotted,mark=diamond*,color=green] coordinates {(0,48.500943183898926) (1,40.144917368888855) (2,38.988474011421204) (3,35.723498463630676) (4,34.35797691345215) (5,34.923672676086426) (6,34.79485213756561) (7,35.093849897384644) (8,35.619884729385376) (9,36.808741092681885) (10,37.26498484611511) (11,35.26272773742676) (12,42.8566575050354)};

\legend {GPT-2 $n=5$, GPT-2 $n=8$, CLIP $n=5$, CLIP $n=8$};

\end{axis}
\end{tikzpicture}
\caption{The five highest-magnitude neuron activations make up more than 97\% of the length of GPT-2 CWEs, compared to 39\% of the length of CLIP CWEs.}
\label{cwe_magnitude}
\end{figure}
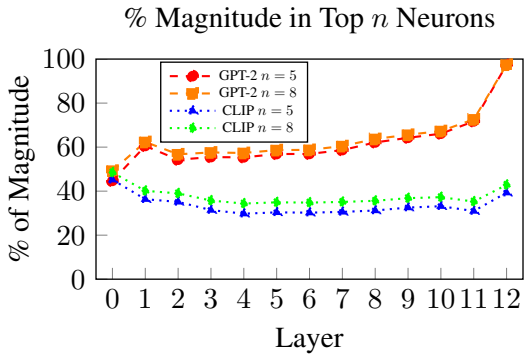

\begin{table*}[t]
\centering
\begin{tabular}
{|l||l|c|l|c|l|c|l|c|l|c|}
 \hline
 \multicolumn{11}{|c|}{Performance by Intrinsic Evaluation Task} \\
 \hline
 {Task} & \multicolumn{2}{c}{RG65} & \multicolumn{2}{|c|}{WS-353} & \multicolumn{2}{|c|}{SL-999} & \multicolumn{2}{|c|}{ValNorm} & \multicolumn{2}{|c|}{SV-3500}\\
 \cline{2-9}
 \hline
 {Layer} &\multicolumn{1}{|c|}{Best} & \multicolumn{1}{|c|}{Top} & \multicolumn{1}{|c|}{Best} & \multicolumn{1}{|c|}{Top} & \multicolumn{1}{|c|}{Best} & \multicolumn{1}{|c|}{Top} & \multicolumn{1}{|c|}{Best} & \multicolumn{1}{|c|}{Top} & \multicolumn{1}{|c|}{Best} & \multicolumn{1}{|c|}{Top}\\
 \hline
   GPT-2 - no BOS & .09 (1) & .01 & .14 (1) & .12 & .05 (5) & .02 & .43 (7) & .25 & .01 (8) & .00\\
 GPT-2 - w/ BOS & .44 (7) & .23 & .44 (9) & .25 & .25 (8) & .11 & .76 (7) & .33 & .21 (8) & .07\\
CLIP & \textbf{.88 (8)} & .70 & \textbf{.72 (6)} & \textbf{.51} & \textbf{.48 (9)} & \textbf{.39} & \textbf{.88 (4)} & .72 & \textbf{.30 (4)} & \textbf{.17}\\
GPT-2 EOS & .32 (12) & .32 & .31 (3) & .10 & .16 (4) & .05 & .61 (6) & .17 & .10 (4) & -.01\\
CLIP EOS & .73 (12) & \textbf{.73} & .49 (5) & .45 & .34 (11) & .34 & .84 (5) & \textbf{.80} & .14 (11) & .13\\
 \hline
\end{tabular}
\caption{CLIP CWEs outperform GPT-2 CWEs on every intrinsic evaluation task examined. The "EOS" token corresponds to the model's sentence embedding. The best layer corresponds to the layer which a representation achieves the highest score for a task. All scores are Spearman's $\rho$, except for ValNorm, which uses Pearson's $\rho$.}
\label{intrinsic_eval_table}
\end{table*}

\section{Results}

CLIP CWEs are less anisotropic than GPT-2 embeddings, and CLIP outperforms GPT-2 on word-level and sentence-level semantic evaluations.

\subsection{Embedding Geometry}

As illustrated in Figure \ref{selfsim_by_layer}, the self-similarity of CWEs is lower in every layer of the CLIP LM than in GPT-2. Self-similarity in both models is at its highest in the top layer, at $.96$ in GPT-2 and $.24$ in the CLIP LM. The self-similarity of CWEs in GPT-2 never falls below $.55$ in any layer, whereas the self-similarity of CWEs in CLIP falls to $.06$ in layer 4. As shown in Figure \ref{cwe_magnitude}, we also find that the five highest-magnitude neuron activations in the top layer of GPT-2 make up more than $97$\% of the magnitude of GPT-2 CWEs, compared to only $39$\% of the magnitude of CLIP CWEs. For both models, there is a small increase (less than $3$ percentage points in each layer) using the $8$ highest neuron activations. Given that \citet{mu2018all} found that high-magnitude dimensions cause high anisotropy and distort semantics in static word embeddings, and that \citet{ethayarajh2019contextual} suggests increasing isotropy to improve CWE representational quality, we would expect that CLIP CWEs would have more semantic geometry than GPT-2 CWEs.

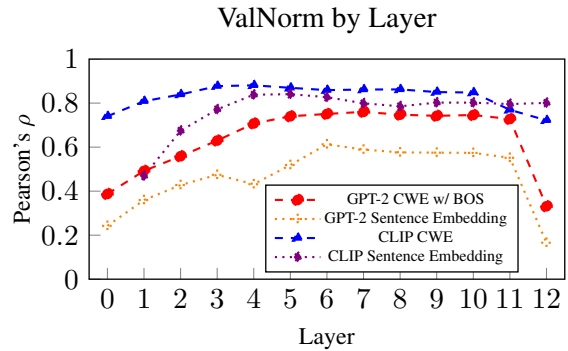
\begin{figure}[htbp]
\begin{tikzpicture}
\begin{axis} [
    height=4.5cm,
    width=.49\textwidth,
    line width = .5pt,
    ymin = 0,
    ymax = 1,
    xmin=-.5,
    xmax=12.5,
    ylabel=Pearson's $\rho$,
    ylabel shift=-5pt,
    xtick = {0,1,2,3,4,5,6,7,8,9,10,11,12},
    label style={font=\small},
    xtick pos=left,
    ytick pos = left,
    title= ValNorm by Layer,
    xlabel= {Layer},
    legend style={at={(.37,.03)},anchor=south west,nodes={scale=0.55, transform shape}}
]

\addplot[thick,dashed,mark=*,color=red] coordinates {(0,0.3871325891628797) (1,0.49206687370054186) (2,0.5585997639857997) (3,0.6298136037538988) (4,0.7074224857412258) (5,0.7400848607830601) (6,0.7507026096563207) (7,0.7602451883929191) (8,0.7473263266865873) (9,0.7422183591889401) (10,0.7457997429727473) (11,0.72771680224806) (12,0.3312864029183524)};

\addplot[thick,dotted,mark=+,color=orange] coordinates {(0,0.2423096432954955) (1,0.3595900803429991) (2,0.4278486563044705) (3,0.4742186284002401) (4,0.43149278386272105) (5,0.5187950454199577) (6,0.6127606533601846) (7,0.5887993798478772) (8,0.5760858077363148) (9,0.5748551663885989) (10,0.5738517683732255) (11,0.5515059567618654) (12,0.16578685085287967)};

\addplot[thick,dashed,mark=triangle*,color=blue] coordinates {(0,0.7401704518199081) (1,0.8084821864344799) (2,0.8392055680127248) (3,0.877009521290169) (4,0.8807525111434503) (5,0.8694304274234743) (6,0.8588839065094855) (7,0.8616635403880518) (8,0.8621365063620802) (9,0.8501115054501778) (10,0.8479462006428529) (11,0.769307659496972) (12,0.7217013533978778)};

\addplot[thick,dotted,mark=diamond*,color=violet] coordinates {(1,0.4691182434071479) (2,0.6743104773759937) (3,0.7713248889877345) (4,0.8380594012768678) (5,0.8405571466243261) (6,0.8269754013154909) (7,0.7989675993772343) (8,0.7853885831867056) (9,0.8023484960341442) (10,0.8020037500320615) (11,0.7956983749601745) (12,0.8008721909234872)};

\legend {GPT-2 CWE w/ BOS, GPT-2 Sentence Embedding, CLIP CWE, CLIP Sentence Embedding};

\end{axis}
\end{tikzpicture}
\caption{CLIP CWEs match the state of the art on the ValNorm intrinsic evaluation task in layer 4.}
\label{cwe_valnorm}
\end{figure}

\subsection{Word-Level Intrinsic Evaluation Tasks}
As shown in Table \ref{intrinsic_eval_table}, CLIP embeddings outperform GPT-2 embeddings on all five of the word-level intrinsic evaluation tasks we study, and non-trivially improve the corpus-based state of the art for the RG65 intrinsic evaluation to Spearman's $\rho = .88$.\footnote{According to the ACL leaderboard at \url{https://aclweb.org/aclwiki/RG-65_Test_Collection_(State_of_the_art)}. Precisely, CLIP embeddings achieve Spearman's $\rho = .876$ on this task.} As visualized in Figure \ref{cwe_valnorm}, CLIP embeddings also match the state of the art for the ValNorm intrinsic evaluation task \cite{toney2020valnorm}, previously achieved by the GloVe embeddings of \citet{pennington2014glove}. For every task except SV-3500, CLIP embeddings outperform the results obtained for GPT-2 by \citet{bommasani2020interpreting}, who create static word embeddings by pooling over CWEs obtained from $100,000$ encoded contexts, both in GPT-2 small and in GPT-2 medium, a 24-layer model which forms $1,024$-dimensional embeddings. For SV-3500, \citet{bommasani2020interpreting} obtain Spearman's $\rho=.31$ in layer 6 of GPT-2 small from embeddings formed using CWEs $100,000$ from contexts.

Our results also indicate that adding the BOS token in GPT-2 significantly improves results on word-level semantic intrinsic evaluation tasks in the decontextualized setting. ValNorm scores improve from .59 to .76 in layer 7, and RG65 scores improve from .01 to .44 in the same layer. On every test, simply adding the BOS token outperforms results reported by \citet{bommasani2020interpreting} on embeddings obtained using the pooling methodology for $10,000$ contexts, both in GPT-2 small and GPT-2 medium \citet{bommasani2020interpreting}. While adding the BOS token does not match the results of applying the pooling method to 50,000 or 100,000 contexts, this marked improvement indicates that using the BOS token is a simple, computationally efficient, and easily replicated way of obtaining static reductions of CWEs, with better quality than representations requiring ten thousand contexts to form.

Finally, we find that CLIP EOS token embeddings outperform CWEs in the top layer on two of five word-level intrinsic evaluation tasks, and nearly equal the performance of CLIP CWEs on the other three tasks. ValNorm scores fall to .72 for CLIP CWEs in the top layer, but increase to $.80$ for CLIP EOS token embeddings in that layer; and RG65 scores fall to $.70$ in the top layer for CLIP CWEs, but increase to $.73$ for CLIP EOS token embeddings. CWEs lose some of their mutual information with the input word as the model forms predictions about the next word in the sequence \cite{voita-etal-2019-bottom}, but our findings indicate that the EOS token must maintain the semantic information of a context in the top layers, such that it can be projected to the joint language-image space and  accurately associated with an image.

Additional visualizations of CLIP and GPT-2 performance on word-level intrinsic evaluation tasks are included in Appendix \ref{appendix_a}.

\begin{figure}[htbp]
\begin{tikzpicture}
\begin{axis} [
    height=4.5cm,
    width=.5\textwidth,
    line width = .5pt,
    ymin = 0,
    ymax = 1,
    xmin=-.5,
    xmax=12.5,
    ylabel=Spearman's $\rho$,
    ylabel shift=-5pt,
    xtick = {0,1,2,3,4,5,6,7,8,9,10,11,12},
    xtick pos=left,
    ytick pos = left,
    title=STS Benchmark Performance,
    xlabel= {Layer},
    legend style={at={(.03,.58)},anchor=south west,nodes={scale=0.7, transform shape}}
]
\addplot[thick,dotted,mark=square*,color=orange] coordinates {(0,0.06703665045831869) (1,0.27021013234233393) (2,0.3406758242961176) (3,0.30656418158158427) (4,0.33764200465085337) (5,0.3322137967370876) (6,0.42736172255637017) (7,0.4493124175040565) (8,0.4417037553519928) (9,0.4467696718970579) (10,0.44739260812885057) (11,0.4403978439340459) (12,0.1453905326704606)};

\addplot[thick,dashed,mark=*,color=red] coordinates {(0,0.0606312371952626) (1,0.2523503461029344) (2,0.22541470054355467) (3,0.26836777063289574) (4,0.2747329029931493) (5,0.28357313539153434) (6,0.27662462802354754) (7,0.2762909256470887) (8,0.2659164496139524) (9,0.25635919804950896) (10,0.25842749942885684) (11,0.26066007545401115) (12,0.13564772485053284)};

\addplot[thick,dashed,mark=triangle*,color=blue] coordinates {(0,0.047506373030760524) (1,0.37338503605197876) (2,0.3583826054144865) (3,0.38115304375926723) (4,0.45336694399945204) (5,0.47783769275303517) (6,0.4390958082117833) (7,0.49122794431907024) (8,0.513329799237829) (9,0.5989569712499385) (10,0.6178891526061417) (11,0.6732074917904058) (12,0.7271757714244256)};

\legend {GPT-2 Sentence Embedding, GPT-2 Last Token, CLIP Sentence Embedding};

\end{axis}
\end{tikzpicture}
\caption{CLIP sentence embeddings outperform GPT-2 embeddings on the STS Benchmark.}
\label{sts_benchmark}
\end{figure}
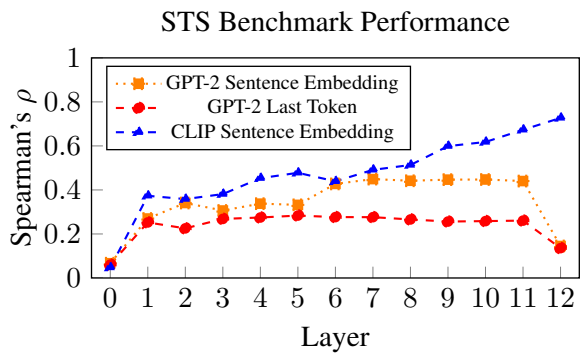

\subsection{Sentence Embeddings}

As shown in Figure \ref{sts_benchmark}, sentence embeddings from the CLIP LM outperform GPT-2 sentence embeddings on the STS benchmark at every layer of the respective models, and the difference in performance grows in the upper layers. CLIP sentence embeddings obtain Spearman's $\rho = .73$ in the top layer, compared to no greater than $.45$ for GPT-2 embeddings. Even using the EOS token, GPT-2 sentence embeddings exhibit properties similar to CWEs in the model, and lose semantic information in the upper layers, while CLIP sentence embeddings improve in semantic quality through the top layer.

\begin{figure}[htbp]
\begin{tikzpicture}
\begin{axis} [
    height=4.5cm,
    width=.5\textwidth,
    line width = .5pt,
    ymin = 0,
    ymax = 1,
    xmin=-.5,
    xmax=12.5,
    ylabel=Self-Similarity,
    ylabel shift=-5pt,
    xtick = {0,1,2,3,4,5,6,7,8,9,10,11,12},
    xtick pos=left,
    ytick pos = left,
    title=Sentence Embedding Self-Similarity,
    xlabel= {Layer},
    legend style={at={(.02,.05)},anchor=south west,nodes={scale=0.7, transform shape}}
]
\addplot[thick,dotted,mark=square*,color=orange] coordinates {(0,0.9730402881418302) (1,0.9910561629053559) (2,0.9913384375498965) (3,0.9927466823795782) (4,0.9889435273054262) (5,0.9931298306198951) (6,0.9906542513520104) (7,0.9923735095186891) (8,0.9956014454453507) (9,0.9966160378780184) (10,0.9980507272030487) (11,0.9989681227883501) (12,0.9997062357207929)};

\addplot[thick,dashed,mark=*,color=red] coordinates {(0,0.8263383449543645) (1,0.8659200880339724) (2,0.8364031607385781) (3,0.818006670656244) (4,0.7880104637654716) (5,0.7585318159847436) (6,0.7355085443500027) (7,0.7180983596530568) (8,0.7080575444689063) (9,0.716388271610792) (10,0.7497098631763114) (11,0.8307736698963495) (12,0.9871713395576461)};

\addplot[thick,dashed,mark=triangle*,color=blue] coordinates {(0,0.9835132972866765) (1,0.9163708351600007) (2,0.9161270651193234) (3,0.8584315020303688) (4,0.7271210548694589) (5,0.6205153348722512) (6,0.5734928255879785) (7,0.5159710128525239) (8,0.530912286421416) (9,0.4904356310983111) (10,0.5461696975033264) (11,0.4929907090586772) (12,0.25140846813935347)};

\legend {GPT-2 Sentence Embedding, GPT-2 Last Token, CLIP Sentence Embedding};

\end{axis}
\end{tikzpicture}
\caption{CLIP sentence embeddings become less self-similar as the layer index increases, while GPT-2 sentence embeddings remain highly anisotropic.}
\label{sentence_similarity}
\end{figure}
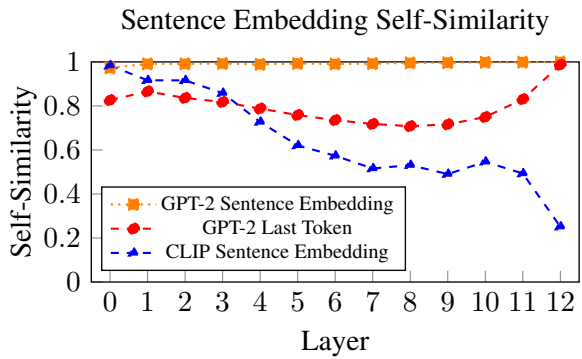

As shown in Figure \ref{sentence_similarity}, CLIP sentence embeddings become increasingly dissimilar as the layer index increases. This is in stark contrast to GPT-2, wherein sentence embeddings using the EOS token have self-similarity $\geq .97$ in every layer, and indicates that the contrastive visual semantic objective of CLIP forces fine-grained differentiation of sentence-level semantics. 

\section{Discussion}

Our findings are straightforward, but it is not obvious that they should occur. The training objective of CLIP is not to produce high-quality CWEs, or even sentence embeddings. Indeed, \citet{radford2021learning} spend little time discussing the CLIP language model, noting that they did not see significant performance improvements by scaling up the size of the model. However, in creating the first broadly accurate zero-shot image classifier, \citet{radford2021learning} have also created a zero-shot sentence encoder which substantially outperforms the version of its underlying architecture trained on language modeling. Moreover, without the need for computationally expensive pooling methodologies, and despite having less than half the parameter count of GPT-2 small, the CLIP LM produces CWEs which match or exceed the best performance ever realized with a corpus-based approach on two intrinsic evaluation tasks, and outperform embeddings formed from $100,000$ encoded contexts in GPT-2 medium \cite{bommasani2020interpreting}.

CLIP embeddings show that the high anisotropy observed by \citet{ethayarajh2019contextual} is not the inevitable result of contextualization, nor even of a specific language modeling architecture, but is connected to the pretraining objective of the model. When trained on a contrastive visual semantic objective, CWEs formed by CLIP have much lower self-similarity at every layer of the model in comparison with GPT-2. This is remarkable because CLIP does not actually project CWEs into the joint language-image space. While we might expect CLIP sentence embeddings, which are projected into the language-image space, to have different properties from the CWEs formed by GPT-2, it does not necessarily also follow that the CWEs formed by CLIP would also be so different from those in GPT-2. Indeed, we still observe the increased self-similarity in the top layer reported by \citet{ethayarajh2019contextual}, and the loss of semantic information related to the input token in the upper layers, as reported by \citet{voita-etal-2019-bottom}. However, these effects are much less pronounced in CLIP than they are in GPT-2, indicating that the contrastive visual semantic objective of the model has regularizing effects that shape more than just the projected sentence embedding.

Our findings suggest that language models trained on visual semantic objectives are likely to privilege the encoding of semantic information, which is essential to matching a caption to an image. The more isotropic representations we observe reflect the objective of the model, which requires differentiating fine-grained semantic information. That models trained on visual semantic objectives would form embeddings to reflect the semantics of a word or sentence more than would a causal language model makes intuitive sense. Through the lens of the training objective, it is more problematic for a causal language model to predict a syntactically invalid continuation of a sentence, such as an incorrect part of speech, than to predict a somewhat unexpected but still syntactically valid continuation of a sentence. When a language model is trained to encode and associate the correct text caption with a matching image, however, the semantic content of the text becomes at least as important as its syntactic properties.

\subsection{Limitations and Future Work}

Our work shows that a pretraining objective which is both visual semantic and contrastive in nature  results in isotropic, highly semantic CWEs and sentence representations, in stark contrast to the representations formed by the same architecture when trained on a language modeling objective. However, further work is needed to address to what extent the results we observe are the result of contrastive training, and to what extent they are the result of visual semantic training. It is possible that a contrastive training objective, wherein the model must discriminate between correct and incorrect options, will result in isotropic and highly semantic embeddings even if both models produce linguistic representations. On the other hand, encoding language for the purpose of performing visual semantic tasks may be particularly important for achieving the effects seen in CLIP, as images lack a grammatical structure and are primarily semantic in composition. Future work might perform a direct assessment between representations obtained from the CLIP LM and representations learned by contrastive text-only models such as those recently introduced by \citet{neelakantan2022text}.

This work examines semantics in contextualized representations without postprocessing, using cosine similarity as the similarity metric. While this is a common experimental design evaluated frequently in prior work, it is not the only way of assessing semantics in contextualized word embeddings. For example, recent work indicates that semantics can be better isolated in language models like GPT-2 by postprocessing and transforming the embedding space using methods such as removing high-magnitude directions with principal component analysis \cite{wolfe2022vast,timkey2021all}.\footnote{CLIP still outperforms GPT-2 in nearly every case over intrinsic evaluation results reported after postprocessing, and CLIP embeddings may also exhibit improvements from comparable manipulations of the embedding space.} Future work might assess whether these postprocessing techniques, or methods which assess semantics using mutual information \cite{voita-etal-2019-bottom} or linear probes \cite{tenney2019bert}, also indicate that contrastive multimodal pretraining magnifies semantics in the embedding space.

Finally, \citet{radford2021learning} note that CLIP was first intended to be a zero-shot caption generator, a design which has since been realized using the SimVLM architecture of \cite{wang2021simvlm}. Analysis of such models, which are not yet available to the research community in a way which would allow analysis of the underlying architecture, may help to answer questions of whether the contrastive objective or the visual semantic setting is more important for regularizing anisotropy and representing semantics. 

\section{Conclusion}
We find that contrastive visual semantic pretraining produces isotropic CWEs which outperform a language model based on the same architecture on semantic evaluations on both the word level and the sentence level. Our findings indicate that incorporating visual semantic objectives with language models may be useful both to regularize the anisotropy in CWEs and to improve the semantic quality of both word and sentence representations.

\section{Ethical Considerations}
While the contrastive visual semantic objective of CLIP produces semantically rich representations of natural language, we caution that the model is also known to encode harmful societal biases. \citet{goh2021multimodal} find that the CLIP image encoder forms representations which reflect biases against communities marginalized based on religion and on immigration status, and \citet{wang2021gender} and \citet{agarwal2021evaluating} report biases of underrepresentation and stereotypical associations which disproportionately affect women. Moreover, \citet{radford2021learning} state that they use frequency-based heuristics to construct the WebImageText corpus on which CLIP trains. Other research on language models has shown that similar techniques can exacerbate biases against marginalized groups, who are often underrepresented in such datasets \cite{wolfe2021low}. Thus, while our findings are promising for the future of visual semantic AI systems, models like CLIP must be studied further to understand how they represent people, and what the ramifications of such representations are for society.

\section*{Acknowledgements}
This material is based on research partially supported by the U.S. National Institute of Standards and Technology (NIST) Grant 60NANB20D212. Any opinions, findings, and conclusions or recommendations expressed in this material are those of the authors and do not necessarily reflect those of NIST.

\bibliography{anthology,custom}
\bibliographystyle{acl_natbib}

\appendix

\section{Intrinsic Evaluation Performance}\label{appendix_a} 

We include visualizations showing the performance of CLIP and GPT-2 embeddings on the intrinsic evaluation tasks discussed in the paper. 

\begin{figure}[H]
\centering
\begin{tikzpicture}
\begin{axis} [
    height=3.5cm,
    width=7cm,
    line width = .5pt,
    ymin = -0.2,
    ymax = 1,
    xmin=-.5,
    xmax=12.5,
    ylabel=Spearman's $\rho$,
    ylabel shift=-7pt,
    xtick = {0,1,2,3,4,5,6,7,8,9,10,11,12},
    xtick pos=left,
    ytick pos = left,
    title=RG65,
    xlabel= {Layer},
    legend style={at={(.7,.95)},anchor=south west,nodes={scale=0.5, transform shape}}
]
\addplot[thick,dotted,mark=square*,color=orange] coordinates {(0,-0.10108263672398302) (1,0.1259697972417024) (2,0.24304478055603268) (3,0.21380335295204278) (4,0.23808379097000454) (5,0.2526607956126512) (6,0.13916996336487872) (7,0.1410057480575059) (8,0.16154905295119087) (9,0.13670039586170168) (10,0.13348777264960415) (11,0.16651004253721904) (12,0.32213261679518435)};

\addplot[thick,dashed,mark=*,color=red] coordinates {(0,0.12704067164573493) (1,0.3118867065295515) (2,0.359464126480139) (3,0.3559455391526036) (4,0.3705881075342727) (5,0.40682300063398513) (6,0.4278689608601752) (7,0.4436479673848992) (8,0.4177284358913775) (9,0.40817798457378135) (10,0.36658871945390636) (11,0.36112507453537307) (12,0.22737504692967933)};

\addplot[thick,dashed,mark=x,color=green] coordinates {(0,-0.016631335132015184) (1,0.08617260765510627) (2,-0.029940774153562163) (3,-0.03597263814362286) (4,-0.008741831869653187) (5,-0.01195445508175073) (6,-0.010927289837066482) (7,0.007430557089205207) (8,-0.02049959573433672) (9,-0.026640732622768084) (10,-0.035295146173724735) (11,-0.04692178256036347) (12,0.014030640150793363)};

\addplot[thick,dashed,mark=triangle*,color=blue] coordinates {(0,-0.14156449581962655) (1,0.10553576524638808) (2,0.19841772886145317) (3,0.26994776813489035) (4,0.362261512678428) (5,0.4248311742854707) (6,0.39165592234013685) (7,0.3779968100438037) (8,0.37471862309268383) (9,0.5195707771728371) (10,0.5132766582266868) (11,0.6354874677644383) (12,0.7274296844535157)};

\addplot[thick,dashed,mark=diamond*,color=violet] coordinates {(0,0.6432021343894072) (1,0.6723998528340489) (2,0.6767707687688755) (3,0.759359225357424) (4,0.8015167095488265) (5,0.8004021259854457) (6,0.7895185453077275) (7,0.8318945752958713) (8,0.8764560632514284) (9,0.8191315007661777) (10,0.7910702204645909) (11,0.7050943040265518) (12,0.696418035895921)};

\legend {GPT-2 Sentence Emb., GPT-2 CWE, GPT-2 CWE - No BOS, CLIP Sentence Emb., CLIP CWE};

\end{axis}
\end{tikzpicture}

\begin{tikzpicture}
\begin{axis} [
    height=3.5cm,
    width=7cm,
    line width = .5pt,
    ymin = -.2,
    ymax = 1,
    xmin=-.5,
    xmax=12.5,
    ylabel=Spearman's $\rho$,
    ylabel shift=-5pt,
    xtick = {0,1,2,3,4,5,6,7,8,9,10,11,12},
    xtick pos=left,
    ytick pos = left,
    title=WS353,
    xlabel= {Layer},
    legend style={at={(.7,.85)},anchor=south west,nodes={scale=0.5, transform shape}}
]
\addplot[thick,dotted,mark=square*,color=orange] coordinates {(0,0.061317625416031335) (1,0.10945784516621802) (2,0.19589895130694843) (3,0.30572453599721144) (4,0.2547483432991974) (5,0.1982452956595394) (6,0.1489376315347941) (7,0.14760266997830443) (8,0.14604647567496432) (9,0.14849098560133797) (10,0.1482561601476176) (11,0.13248576095331804) (12,0.09601442048349468)};

\addplot[thick,dashed,mark=*,color=red] coordinates {(0,0.12875730028862328) (1,0.1981581313062282) (2,0.3094117654719447) (3,0.3346585998510474) (4,0.36844126611246714) (5,0.38351237838154245) (6,0.40524576636277976) (7,0.42962955087740695) (8,0.4287310350788769) (9,0.43708817321653387) (10,0.43477784042775214) (11,0.39896406740451174) (12,0.24830991175477812)};

\addplot[thick,dashed,mark=x,color=green] coordinates {(0,0.11760039947223075) (1,0.13747673658339218) (2,0.13396383581528132) (3,0.08789555712951092) (4,0.0765695555869972) (5,0.08069269166309281) (6,0.07229364411687195) (7,0.06809949769306486) (8,0.07519427852901116) (9,0.0852345829465127) (10,0.09292047591612242) (11,0.10000144593281447) (12,0.11652141663233805)};

\addplot[thick,dashed,mark=triangle*,color=blue] coordinates {(0,-0.00653878537808514) (1,0.10021513500609092) (2,0.18336201643351485) (3,0.35096051643925713) (4,0.4906683330895627) (5,0.49160040511800607) (6,0.4180714757838811) (7,0.37909154970163317) (8,0.3567077710528095) (9,0.4184671937634675) (10,0.4378135879753896) (11,0.42420053481827547) (12,0.44644231332344186)};

\addplot[thick,dashed,mark=diamond*,color=violet] coordinates {(0,0.4701566101290688) (1,0.6247166051093818) (2,0.6374093172777784) (3,0.6500737931063699) (4,0.6746302694383556) (5,0.7013676278998233) (6,0.7196119864147172) (7,0.70977510038235) (8,0.7001612405122114) (9,0.6982102049168908) (10,0.6919353263305564) (11,0.5847305377594917) (12,0.5101432231087135)};

\legend {GPT-2 Sentence Emb., GPT-2 CWE, GPT-2 CWE - No BOS, CLIP Sentence Emb., CLIP CWE};

\end{axis}
\end{tikzpicture}

\begin{tikzpicture}
\begin{axis} [
    height=3.5cm,
    width=7cm,
    line width = .5pt,
    ymin = -.3,
    ymax = 1,
    xmin=-.5,
    xmax=12.5,
    ylabel=Spearman's $\rho$,
    ylabel shift=-5pt,
    xtick = {0,1,2,3,4,5,6,7,8,9,10,11,12},
    xtick pos=left,
    ytick pos = left,
    title=SL-999,
    xlabel= {Layer},
    legend style={at={(.7,.75)},anchor=south west,nodes={scale=0.5, transform shape}}
]
\addplot[thick,dotted,mark=square*,color=orange] coordinates {(0,0.060255032559977266) (1,-0.0008973144012544435) (2,0.04805284509462224) (3,0.1413246580055671) (4,0.16132422174519337) (5,0.0988755412768235) (6,0.15266740210202234) (7,0.14673182385604033) (8,0.13880769744868873) (9,0.13923587550093133) (10,0.1321309398263217) (11,0.12117405144171665) (12,0.04969475747299957)};

\addplot[thick,dashed,mark=*,color=red] coordinates {(0,0.028227332504461074) (1,0.1281696074370349) (2,0.1669258205337266) (3,0.13971079585633883) (4,0.15724411842314387) (5,0.17922749179140696) (6,0.20542406586349699) (7,0.23416996880636406) (8,0.24659749079863488) (9,0.23492292825940772) (10,0.23156042839918098) (11,0.2254597690506792) (12,0.10662508511386146)};

\addplot[thick,dashed,mark=x,color=green] coordinates {(0,0.018208303712010446) (1,0.029901912583012922) (2,0.03613676462783549) (3,0.03618171582215604) (4,0.044460928419074765) (5,0.046428579462932495) (6,0.04059367410729652) (7,0.03915173150208802) (8,0.04032966298986533) (9,0.04469240363109877) (10,0.04260384392330314) (11,0.029447434968954706) (12,0.016574144767104684)};

\addplot[thick,dashed,mark=triangle*,color=blue] coordinates {(0,-0.0772348752468519) (1,-0.2181324974944059) (2,-0.15452093809669945) (3,-0.04186063987246725) (4,0.15008944725309015) (5,0.1935499235959505) (6,0.2028860337546971) (7,0.18036591410867422) (8,0.20106402845034743) (9,0.2570420228239757) (10,0.28623427880856395) (11,0.33871151848679726) (12,0.33512696112828844)};

\addplot[thick,dashed,mark=diamond*,color=violet] coordinates {(0,0.29060407765220336) (1,0.38055082102769755) (2,0.37529095665683654) (3,0.41878810155455165) (4,0.4407845944779737) (5,0.448607105168919) (6,0.47036782524258774) (7,0.4715333653192799) (8,0.4751269920216532) (9,0.47657867681813554) (10,0.4717444938670259) (11,0.38922627131202253) (12,0.3900285321099923)};

\legend {GPT-2 Sentence Emb., GPT-2 CWE, GPT-2 CWE - No BOS, CLIP Sentence Emb., CLIP CWE};

\end{axis}
\end{tikzpicture}

\begin{tikzpicture}
\begin{axis} [
    height=3.5cm,
    width=7cm,
    line width = .5pt,
    ymin = -.2,
    ymax = 1,
    xmin=-.5,
    xmax=12.5,
    ylabel=Spearman's $\rho$,
    ylabel shift=-5pt,
    xtick = {0,1,2,3,4,5,6,7,8,9,10,11,12},
    xtick pos=left,
    ytick pos = left,
    title=SV-3500,
    xlabel= {Layer},
    legend style={at={(.7,.75)},anchor=south west,nodes={scale=0.5, transform shape}}
]
\addplot[thick,dotted,mark=square*,color=orange] coordinates {(0,-0.013931264781611854) (1,0.00014408015037596322) (2,0.013613597945063517) (3,0.07071996242404753) (4,0.09567027810251051) (5,0.09285020499238487) (6,0.07989202745012676) (7,0.07562104511880306) (8,0.07974260087414768) (9,0.08738069730469364) (10,0.0860786033757594) (11,0.08889746742338082) (12,-0.010200570657959655)};

\addplot[thick,dashed,mark=*,color=red] coordinates {(0,0.0008863543984573186) (1,0.07670322733128614) (2,0.11786251633824045) (3,0.14088635800220442) (4,0.1670543238835146) (5,0.17179278631815761) (6,0.18794930391003148) (7,0.1994301366858725) (8,0.2093714004332078) (9,0.2017695149904309) (10,0.19247502645826225) (11,0.18085210280267983) (12,0.07343996626805843)};

\addplot[thick,dashed,mark=x,color=green] coordinates {(0,-0.02605992972779704) (1,-0.014469835895376837) (2,-0.012169020736389247) (3,0.00040573501590993917) (4,0.010314772553983602) (5,0.007563088330032952) (6,0.010425749587119523) (7,0.011916098680902555) (8,0.012870780976957083) (9,0.01194665008142878) (10,0.00676700376545592) (11,0.008099179001252574) (12,-0.003825116500275881)};

\addplot[thick,dashed,mark=triangle*,color=blue] coordinates {(0,-0.023501541419921352) (1,-0.11219138491441237) (2,-0.08812264744409776) (3,0.004352788312977127) (4,0.10038869900047141) (5,0.10847001089723202) (6,0.09727538180680928) (7,0.0790870633706029) (8,0.08094773072204495) (9,0.12117748845634199) (10,0.1371760322665687) (11,0.14454851531847795) (12,0.13218056129476394)};

\addplot[thick,dashed,mark=diamond*,color=violet] coordinates {(0,0.14757097328428467) (1,0.22299777793040568) (2,0.24670680084249802) (3,0.27919247201114067) (4,0.30538447983822004) (5,0.30214491242870056) (6,0.300701562920923) (7,0.2965095785735271) (8,0.2817597944562677) (9,0.2585047216551069) (10,0.23039913809613263) (11,0.16776134419229943) (12,0.16658355310064918)};

\legend {GPT-2 Sentence Emb., GPT-2 CWE, GPT-2 CWE - No BOS, CLIP Sentence Emb., CLIP CWE};

\end{axis}
\end{tikzpicture}

\caption{CLIP CWEs outperform other representations in almost every layer across four intrinsic evaluations, including achieving corpus-based state of the art on RG65 in layer 8, with Spearman's $\rho=.876$..}
\end{figure}
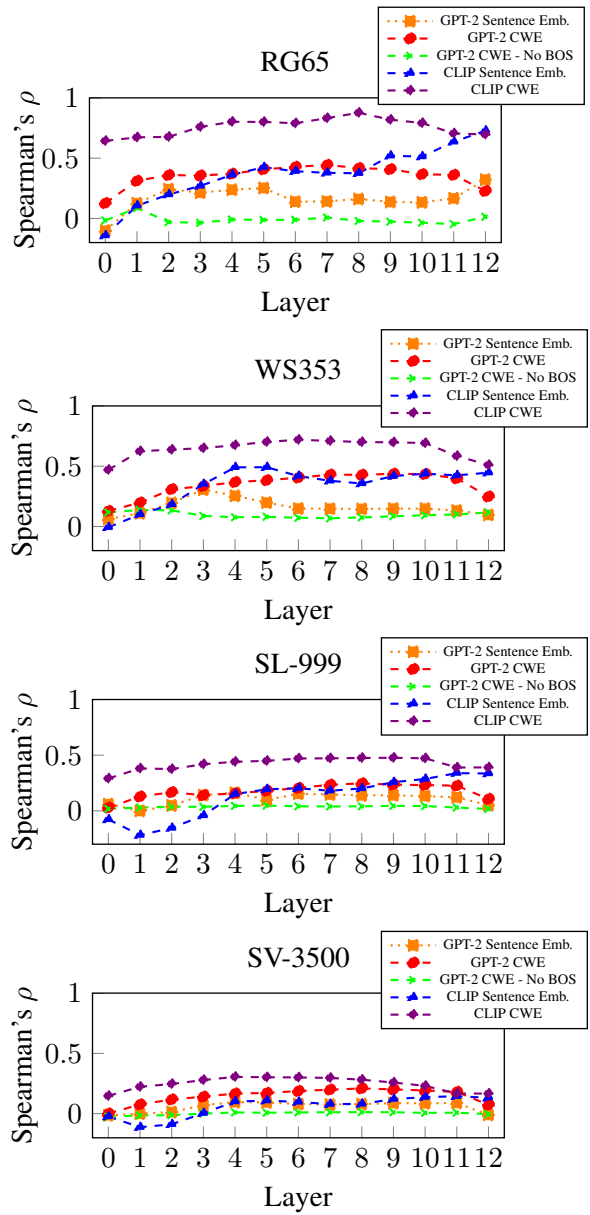

\end{document}